\documentclass[10pt, conference]{ieeeconf}      

\IEEEoverridecommandlockouts                              

\newtheorem{theorem}{Theorem}[section]

\usepackage[ruled,vlined]{algorithm2e}
\usepackage{url}

\newcommand{\qed}{\nobreak \ifvmode \relax \else
      \ifdim\lastskip<1.5em \hskip-\lastskip
      \hskip1.5em plus0em minus0.5em \fi \nobreak
      \vrule height0.75em width0.5em depth0.25em\fi}

\usepackage{amsmath,amssymb}

\usepackage{tabularx}
\usepackage{tikz,hyperref,graphicx,units}
\usepackage{subfigure}
\usepackage{benktools}
\usepackage{bbm}
\renewcommand{\baselinestretch}{.5}

\usepackage{caption}
\usepackage{epstopdf}

\usepackage{sidecap,wrapfig}
\usepackage[ruled,vlined]{algorithm2e}
\DeclareMathOperator*{\argmin}{arg\,min}

\newcommand{\bu}{\mathbf{u}}

\newcommand{\bx}{\mathbf{x}}

\newcommand{\specialcell}[2][c]{ \begin{tabular}[#1]{@{}c@{}}#2\end{tabular}}

\newcommand{\ns}{HC }
\newcommand{\nc}{RC }

\newcolumntype{L}[1]{>{\RaggedRight\hspace{0pt}}p{#1}}
\newcolumntype{R}[1]{>{\RaggedLeft\hspace{0pt}}p{#1}}

\newboolean{include-notes}
\setboolean{include-notes}{true}
\newcommand{\adnote}[1]{\ifthenelse{ \boolean{include-notes}}%
 {\textcolor{blue}{\textbf{#1}}}{}}
 
 \newcommand{\sknote}[1]{\ifthenelse{ \boolean{include-notes}}%
 {\textcolor{blue}{\textbf{SK: #1}}}{}}
 
  \newcommand{\mlnote}[1]{\ifthenelse{ \boolean{include-notes}}%
 {\textcolor{purple}{\textbf{ML: #1}}}{}}
 
 \newcommand{\jmnote}[1]{\ifthenelse{ \boolean{include-notes}}%
 {\textcolor{orange}{\textbf{JM: #1}}}{}}

\renewcommand{\baselinestretch}{0.95}
\usepackage{times}
\usepackage{microtype}

\title{Comparing Human-Centric and Robot-Centric \\
Sampling for Robot Deep Learning from Demonstrations}

\author{Michael Laskey$^1$, Caleb Chuck$^1$, Jonathan Lee$^1$, Jeffrey Mahler$^1$,\\ Sanjay Krishnan$^1$, Kevin Jamieson$^1$, Anca Dragan$^1$, Ken Goldberg$^{1,2}$
\thanks{$^1$ Department of Electrical Engineering and Computer Sciences; {\small \{mdlaskey, calebchuck, jonathan\textunderscore lee, jmahler,sanjaykrishnan, , anca\}@berkeley.edu} }%
\thanks{$^2$ Department of Industrial Engineering and Operations Research; {\small goldberg@berkeley.edu}}%
\thanks{$^{1-2}$ University of California, Berkeley;  Berkeley, CA 94720, USA}%
}

\begin{document}

\maketitle
\thispagestyle{empty}
\pagestyle{empty}



\begin{abstract}
Motivated by recent advances in Deep Learning
for robot control, this paper considers two learning algorithms
in terms of how they acquire demonstrations from fallible human supervisors. “Human-Centric”
(HC) sampling is a standard supervised learning algorithm,
where a human supervisor demonstrates the task by teleoperating
the robot to provide trajectories consisting of state-control
pairs. “Robot-Centric” (RC) sampling is an increasingly
popular alternative used in algorithms such as DAgger, where a
human supervisor observes the robot execute a learned policy
and provides corrective control labels for each state visited.
We suggest RC sampling can be challenging for human supervisors and
prone to mislabeling. RC sampling can also induce error in
policy performance because it repeatedly visits areas of the
state space that are harder to learn. Although policies learned
with RC sampling can be superior to HC sampling for standard
learning models such as linear SVMs, policies learned with HC
sampling may be comparable to RC when applied to expressive learning models such as deep learning and
hyper-parametric decision trees, which can achieve very low
error provided there is enough data. We compare HC and RC using a grid world environment
and a physical robot singulation task. In the latter the input is
a binary image of objects on a planar worksurface and the
policy generates a motion in the gripper to separate one object
from the rest. We observe in simulation that for linear SVMs,
policies learned with RC outperformed those learned with HC
but that using deep models this advantage disappears. We also
find that with RC, the corrective control labels provided by
humans can be highly inconsistent. We prove there exists a class
of examples in which at the limit, HC is guaranteed to converge to
an optimal policy while RC may fail to converge. These results
suggest a form of HC sampling may be preferable for highly-expressive
learning models and human supervisors.

 \end{abstract}

\section{Introduction} 
Learning from Demonstrations (LfD) is a well-established
approach where a robot learns a policy from example
trajectories provided by a fallible human supervisor~\cite{laskeyrobot, argall2009survey,pomerleau1989alvinn}. In principle, learning a state-feedback policy is
a supervised learning problem: regression from observed
states to observed controls from the set of example trajectories.
However, if the robot cannot exactly replicate
the policy of the supervisor, such as due to modeling
errors or insufficient data the training error
achieved on the example trajectories may not be indicative
of the actual execution error when using the policy since
the robot could visit a very different distribution of states.
Algorithms such as DAgger have been proposed to address
this problem: after an initial policy is learned, the supervisor observes the robot executing a learned policy and
retro-actively provides corrective control labels for each state
visited.

A recent trend in Machine Learning is to use more
expressive models such Deep Neural Networks and Decision
Trees~\cite{levine2015end}. Given enough training data, such models can represent highly complex feedback policies and may help alleviate the discrepancy between training and execution
error. This paper studies the relative merits of DAgger-like
methods incorporating a human supervisor, and in the context of
model expressiveness.

We consider two classes of learning algorithms that differ
in how they acquire demonstrations. “Human-Centric” (HC)
sampling is the standard supervised learning algorithm, where
a human supervisor demonstrates the task by tele-operating
the robot to provide trajectories consisting of state-action
pairs~\cite{argall2009survey}. “Robot-Centric” (RC) sampling is an increasingly popular strategy used in algorithms such as DAgger, where
a human supervisor observes the robot executing a learned
policy and retro-actively provides corrective control labels for
each state visited~\cite{ross2010efficient,ross2010reduction,ross2013learning}. The intuition behind this is to provide examples to the robot in states it is likely to visit.

\begin{figure}
\center
\includegraphics[width=0.5\textwidth]{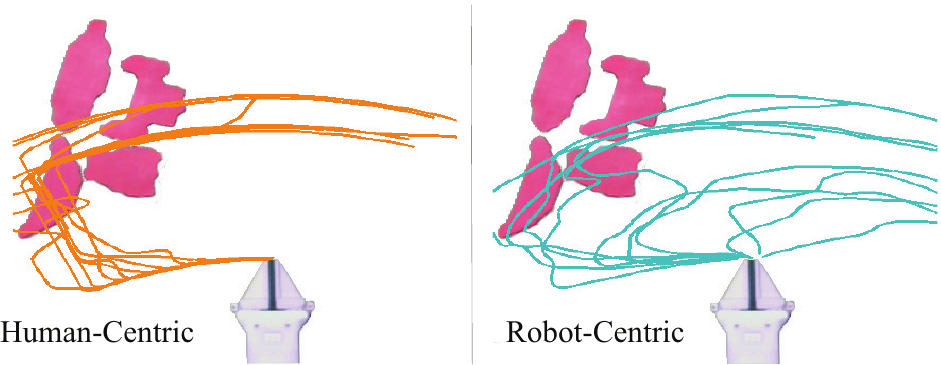}
\caption{
    \footnotesize
HC and RC trajectories used to learn a robot singulation task: top view of 4 objects on a planar worksurface.  Left: In HC, 10 trajectories where a human demonstrates the task by tele-operating a 2-DOF robot to separate one object from a connected set of objects. Right: In RC, after initial training, 10 trajectories of the highly-suboptimal robot policy are executed and a  human provides corrective control labels for each.  Note that many of the RC trajectories spend considerable time in areas of the workspace that will not be useful after the task is learned.}
\vspace*{-20pt}
\label{fig:teaser}
\end{figure}

While RC methods have significant advantages, they can come at a cost for the human supervisor. In many RC methods, a human
retroactively provides corrective feedback to the robot
without observing the effect of their suggested control. This
can require additional cognitive effort on the human supervisor
who must predict how the world behaves without observing the outcome
of their control.

A second cost of RC sampling is that it collects labels on
states that occur due to previous errors. These states may require complex corrective actions that do not occur when
the supervisor manually performs the task. For example, consider the
task of singulating (i.e. separate) an object from a set of
objects on a table (Fig. 1). Demonstrations involve moving
forward and side to side to push obstacles out of the way.
However, if the robot makes errors and ends up in a failure
state (Fig. 1(right)), then more complex maneuvers will be
required. One potential outcome of this effect is the robots
policy could incur larger learning error and fail to converge to the
supervisor`s policy.

The key question is whether the benefits of RC outweigh
these practical challenges, especially if the robot’s policy is an expressive
model that can significantly reduce training error. First, we
show in simulated examples that when the model is highly expressive, HC and
RC are essentially at parity. By varying the expressiveness
of the robots policy, we find that on 100 randomly generated grid
world environments the performance gap between the
two methods diminishes as the expressiveness is increased.
Next, using a point mass control example, we find that RC
sampling can fail to converge in cases when HC converges.

Finally, we illustrate the human factors challenge of
implementing RC with a 10 person pilot study. In this
experiment, each human used RC and HC sampling to
train a Zymark robot to singulate an object. We observed a
statistically significant gap in the average performance of the
policies trained by the two approaches. For 60 trajectories, RC
had a success rate of 40\% compared with 60\%, for HC. Our
post analysis suggests participants had difficulty providing
the retroactive feedback needed for RC sampling.

We conclude the paper with an initial theoretical analysis
that attempts to explain our findings. First, we illustrate a
counter-example wherein as the number of demonstrations
increases, HC converges to an optimum with probability 1,
while for the same model, RC has a non-zero probability
of converging to a suboptimal solution. We emphasize that
this result is not a contradiction of prior theory~\cite{ross2010reduction} as RC
is only theoretically guaranteed in terms of hindsight regret.

\section{Related Work}
Below we summarize related work in HC and RC LfD and their theoretical insights. 

\noindent \textbf {\ns}  Pormeleau et al. used \ns to train a neural network to drive a car on the highway via demonstrations provided by a supervisor. To reduce the number of examples needed to perform a task, they synthetically generated images of the road and control examples~\cite{pomerleau1989alvinn}.  A similar approach was used by Dilmann et al. to teach a robot to perform peg-in-hole insertion using a neural network controller~\cite{dillmann1995acquisition}.  A survey of HC LfD techniques can be found here~\cite{argall2009survey}. 

Ross et al. examined \ns and derived that the error in the worst case from this approach can go quadratic in the time horizon, $T$~\cite{ross2010efficient}. The intuition behind this analysis is that if the distribution induced by the robot's policy is different from the supervisor's, the robot could incur the maximum error. Expressive policies can help manage this term, $T$ , by achieving very low training error. 

\noindent \textbf{\nc}
\nc has been used in numerous robotic examples,
including flying a quadcopter through a forest where the state
space is image data taken from an onboard sensor~\cite{ross2013learning}. Other
successful examples include teaching a robotic wheelchair
to navigate to target positions~\cite{kim2013maximum}, teaching a robot to follow
verbal instructions to navigate across an office building~\cite{duvallet2013imitation}
and teaching a robot to grasp in clutter~\cite{laskeyrobot}.

Ross et al.~\cite{ross2010reduction} analyzed RC sampling as online optimization.
They propose DAgger, an RC sampling algorithm, and show
that the error for the robot’s policy is linear in $T$ for strongly
convex losses (e.g. regularized linear or kernelized regression).
However, the error also depends on the expected loss on the
data collected during RC, which may be high due to observing
complex recovery behaviors. We analyze this effect and show
how it can prevent RC from converging to the supervisor's
policy in Section V-A.

Levine et al. identified that the RC sampling can force
the robot into states that are harder to learn and proposed
weighting the samples to correct for this. They also proposed
forcing the supervisor to guide the robot’s policy to better
regions of the state space~\cite{levine2013variational}. This approach is used for
an algorithmic supervisor and assumes the supervisor can
both be modified and the noise distribution is known. Guided Policy Search also assumes low dimensional state space, where dynamics are known or can be reasonably estimated. We
are interested in maintaining the original assumption of
Ross et al. ~\cite{ross2010reduction}, in which there is an unknown supervisor
whose demonstrations cannot be modified and potentially high dimensional image state spaces.  Finally, He et al. proposed
changing the supervisor’s example to be easier to learn for
the robot’s policy~\cite{he2012imitation}. We are interested in examining how changing the robot’s policy expressiveness can change relative performance of RC and HC.

\textbf{LfD Interfaces:}
Standard techniques for providing demonstrations to a robot are teleoperation, kinesthetic and waypoint specification~\cite{akgun2012keyframe,akgun2012novel,argall2009survey}. Kinesthetic teaching is defined as moving the robot body via a human exerting force on the robot itself. Teleoperation uses an interface such as a joystick or video game controller to control the position of the robot end effector. Waypoint specification, or Keyframes, has a human select positions in the space the robot needs to visit. These methods are forms of HC sampling because the human guides the robot through the task.  We look specifically at teleoperation and compare it to RC's form of retroactive feedback. 

\section{Problem Statement and Background}\label{sec:PS}
The objective of LfD is to learn a policy that matches that of the supervisor on a specified task that demonstrations are collected on.

\noindent\textbf{Modeling Choices and Assumptions:}  We model the system dynamics as Markovian, stochastic, and stationary. Stationary dynamics occur when, given a state and a control, the probability of the next state does not change over time. 

We model the initial state as sampled from a distribution over the state space.
We assume a known state space and set of controls. We also assume access to a robot or simulator, such that we  can sample from the state sequences induced by a sequence of controls.
Lastly, we assume access to a supervisor who can, given a state, provide a control signal label.
We additionally assume the supervisor can be noisy. 

\noindent\textbf{Policies and State Densities.}
Following conventions from control theory, we denote by $\mathcal{X}$ the set consisting of observable states for a robot task, such as images from a camera, or robot joint angles and object poses in the environment.
We furthermore consider a set $\mathcal{U}$ of allowable control inputs for the robot, which can be discrete or
continuous. We model dynamics as Markovian, such that the probability of state $\mathbf{x_{t+1}}\in
\mathcal{X}$ can be determined from the previous state $\mathbf{x}_t\in\mathcal{X}$ and control input $\mathbf{u}_t\in
\mathcal{U}$: 
$$p(\bx_{t+1}|\bu_{t},\bx_{t}, \ldots, \bu_{0}, \bx_{0})=p(\bx_{t+1}|\bu_{t}, \bx_t)$$
We assume a probability density over initial states $p(\bx_0)$.
The environment of a task is thus defined as a specific instance of a control and state space, initial state distribution, and dynamics.


Given a time horizon $T\in \mathbb{N}$, a trajectory $\tau$ is a finite sequence of $T+1$ pairs of states visited and corresponding
control inputs at these states, $\tau = ((\mathbf{x}_0,\mathbf{u}_0), ...., (\mathbf{x}_T,\mathbf{u}_T))$, where $\bx_t\in \mathcal{X}$
and $\bu_t\in \mathcal{U}$ for $t\in \{0, \ldots, T\}$.

A policy is a measurable function $\pi: \mathcal{X} \to \mathcal{U}$ from states to control inputs. 
We consider policies $\pi_{\theta}:\mathcal{X}\to \mathcal{U}$ parametrized by some $\theta\in \Theta$. Under our assumptions, any such policy $\pi_{\theta}$ induces a probability density over the set of  trajectories of length $T+1$: $$p(\tau | \theta)=
p(\bx_0)\prod_{i=0}^{T-1}p(\bx_{t+1}|\bu_t,\bx_t)p(\bu_t|\bx_t,\theta)$$

While we do not assume knowledge of the distributions corresponding to: $p(\bx_{t+1}|\bx_t,\bu_t)$, $p(\bx_0)$ or $p(\bx_t|
\theta)$, we assume that we have a stochastic real robot or a simulator such that for any state
$\bx_t$ and control $\bu_t$, we can observe a sample $\bx_{t+1}$ from the density $p(\bx_{t+1}|\bu_t,\bx_t)$. 
Therefore, when `rolling out' trajectories under a policy
$\pi_{\theta}$, we utilize the robot or a simulator to sample the resulting stochastic trajectories rather than
estimating $p(\tau|\theta)$ itself.

\noindent\textbf{Objective:} We assume access to 
a supervisor, $\pi_{\theta^*}$, where $\theta^*$ may not be contained in $\Theta$. A supervisor is chosen that can achieve a desired level of performance on the task.

We measure the difference between controls using a surrogate loss $l : \mathcal{U} \times \mathcal{U} \rightarrow \mathbb{G}$~\cite{ross2010reduction,ross2010efficient}, where $\mathbb{G} \subset \mathbb{R}^+$.
The surrogate loss can be either an indicator function as in classification or a continuous measure on the sufficient statistics of $p(\bu|\bx,\theta)$.
We measure total loss along a trajectory with respect to the supervisor's policy $\pi_{\theta^*}$ by $J(\theta, \tau) = \sum^T_{t=1} l(\pi_{\theta}(\bx_{t}),\pi_{\theta^*}(\bx_{t}))$.

The obective is to minimize the expected surrogate loss on the distribution of states of the robot: 

\begin{equation}\label{eq:main_obj}
\underset{\theta}{\mbox{argmin }} E_{p(\tau|\theta)} J(\theta, \tau).
\end{equation}

The above objective is challenging to solve because of the coupling between the surrogate loss and the distribution the states the robot is likely to visit. Thus, two classes of algorithms have been proposed (HC and RC). 

\noindent \textbf{HC} In HC, the supervisor provides the robot a set of $N$ demonstration trajectories $\lbrace \tau^1,...,\tau^N \rbrace$ sampled from $p(\tau | \theta^*)$.
This induces a training data set $\mathcal{D}$ of all state-control input pairs from the demonstrated trajectories.
The goal is to find the $\theta^N$ that minimizes the empirical risk, or sample estimate of the expectation.

\begin{equation}\label{eq:main_obj}
\theta^N = \underset{\theta}{\mbox{argmin }} \sum \limits_{i=1}^N J(\theta, \tau_i).
\end{equation}

\noindent \textbf{RC}
Due to sample approximation and learning error, one potential issue with \nc sampling is that $\theta^N$ differs from $\theta^*$.
Therefore it is possible that new states will be visited under $\theta^N$ that would never have been visited under $\theta^*$.
To account for this, prior work has proposed iterative solutions~\cite{ross2010reduction} that attempt to solve this problem by aggregating data on the state distribution induced by the current robot's policy.

Instead of  minimizing the surrogate loss in Eq. \ref{eq:main_obj},   LfD with RC sampling~\cite{ross2010reduction,he2012imitation} attempts to approximate the state distribution the final policy will converge to and minimize the surrogate loss on this distribution.
One popular RC LFD algorithm is DAgger~\cite{ross2010reduction}, which iterates two steps:

\subsubsection{Step 1}
The first step of any iteration $k$ is to compute a $\theta_k$ that minimizes surrogate loss on the current dataset $\mathcal{D}_k=\{(x_i,u_i)|i\in\{1,\ldots,N\}\}$ of demonstrated state-control pairs (initially just the set $\mathcal{D}$ of initial trajectory demonstrations):

 \vspace{-1ex}
\begin{align}\label{eq:super_objj}
\theta_{k} = \underset{\theta}{\argmin} \: \sum_{i=1}^{N} \sum_{t=1}^T  l(\pi_{\theta}(\bx_{i,t}),\bu_{i,t}).
\end{align}

\noindent Note that equal weight is given to each example regardless of how likely it is under the current policy.

 \subsubsection{Step 2}
In the second step of iteration $k$, DAgger rolls out the current policy, $\pi_{\theta_{k}}$, to sample states that are likely under $p(\tau|\theta_{k})$.  For every state visited, DAgger requests the supervisor to provide the appropriate control/label. Formally, for a given sampled trajectory  $\tau = (\bx_0,\bu_0,...,\bx_T,\bu_T )$, the supervisor provides labels $\tilde{\bu}_t$, where $\tilde{\bu}_t \sim \tilde{\pi}(\bx_t) + \epsilon$, where $\epsilon$ is a  zero mean noise term, for $t\in \{0, \ldots, T\}$.
The states and labeled controls are then aggregated into the next data set of demonstrations $\mathcal{D}_{k+1}$:
$$D_{k+1}=\mathcal{D}_k \cup \{(\bx_t,\tilde{\bu_t})|t\in\{0,\ldots,T\}\} $$

\noindent Steps 1 and 2 are repeated for $K$ iterations or until the robot has achieved sufficient performance on the
task.\footnote{In the original DAgger the policy rollout was stochastically mixed with the supervisor, thus with
    probability $\beta$ it would either take the supervisor's action or the robot's. However in \cite{ross2013learning}, it was recommended to display slowed down videos of the robot's execution to the human, which prohibits stochastic mixing.}

\section{Empirical Analysis}

\begin{figure*}
\includegraphics{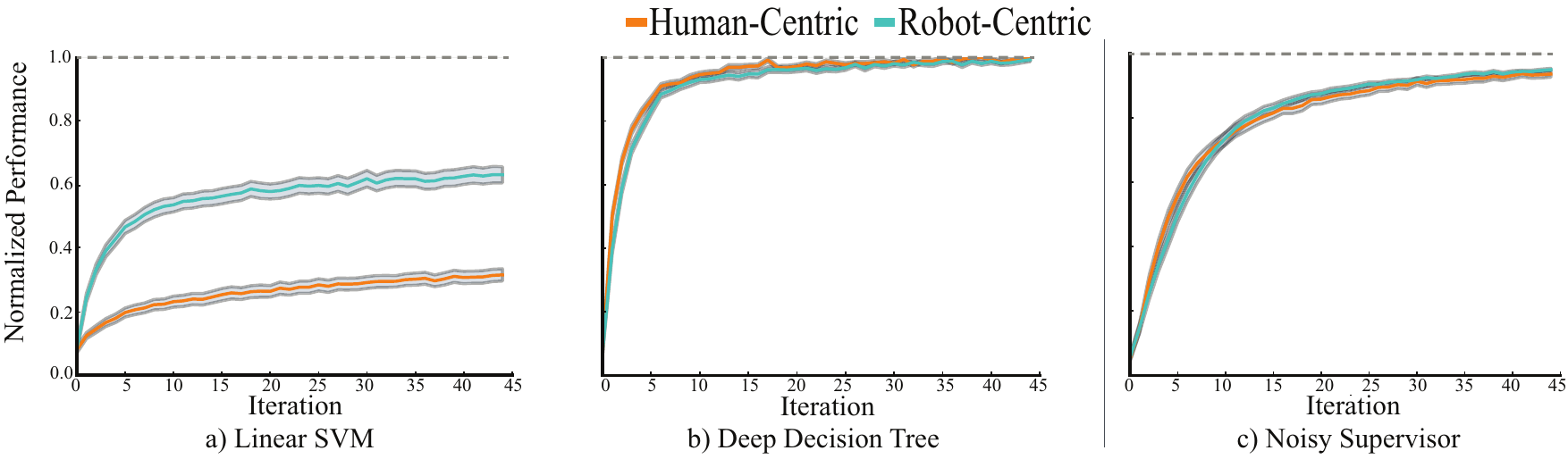}
\caption{
    \footnotesize
We compare RC and HC LfD for low- and high-expressiveness policy classes (a and b respectively) over 100 randomly generated 2D grid world environments, as a function of the amount of data provided to them. RC outperforms in the low-expressive condition, but the performance gap is negligible in the high-expressive condition, when the policy class contains the expected supervisor policy. We also examine the case of noisy supervisor labels (c), in which both techniques take more data to converge, but again perform similarly. The error bars shown are standard error on the mean. 
}
\vspace*{-20pt}
\label{fig:var}
\end{figure*}

We first provide an empirical comparison of HC and RC LfD.
We start with experiments in a Grid World environment, which enable us to vary the robot's policy class over a large number of randomly generated environments. Then, we use a linear dynamical system to exemplify a potential limitation of RC to match the supervisor’s performance when it must learn complex recovery actions.
Finally, we compare HC and RC on a real task. We perform a pilot study with 10 participants, who try to teach a robot how to singulate, or separate an object from clutter. We used DAgger as the example of an RC method in these experiments. 

\subsection{Policy Expressiveness}\label{sec:gdw}
In this experiment, we hypothesize that on random problems with a simulated, perfect supervisor, the performance gap of RC diminishes as the robot's policy becomes more expressive. 

In grid world, we have a robot that is trying to reach a goal state, whereat it receives $+10$ reward. The robot receives $-10$ reward if it touches a penalty state. The robot must learn to be robust to the noise in the dynamics, reach the goal state and then stay there.
The robot has a state space of $(x,y)$ coordinates and a set of actions consisting of $\lbrace$Left, Right, Forward, Backward, Stay$\rbrace$. The grid size for the environment is $15 \times 15$. $8\%$ of randomly drawn states are marked as a penalties, while only one is a goal state. For the transition dynamics, $p(\bx_{t+1}|\bx_{t},\bu_t)$, the robot goes to an adjacent state different from the one desired uniformly at random with probability $0.16$. The time horizon for the policy is $T=30$. 

We use Value Iteration to compute an optimal supervisor. In all settings, we provided RC with one initial demonstration from HC sampling before iteratively rolling out its policy. This initial demonstration set from HC sampling is common in RC methods like DAgger~\cite{ross2010reduction}.

We run all trials over 100 randomly generated environments.
We measure normalized performance, where $1.0$ represents the expected cumulative reward of the optimal supervisor.

\noindent \textbf{Low Expressiveness:} Fig. \ref{fig:var}(a) shows a case when the robot's policy class is empirically not expressive enough to represent the supervisor's policy.
We used a Linear SVM for the policy class representation, which is commonly used in RC ~\cite{ross2010efficient,ross2010reduction,ross2013learning}. RC LfD outperforms HC , which is consistent with prior literature~\cite{ross2010efficient,ross2010reduction}.
This outcome suggests that when the robot is not able to learn the supervisor's policy, RC has a large advantage. Note that neither method converges to the supervisor's performance. 

\noindent \textbf{High Expressiveness:}
We next consider the situation where the robot's policy class is more expressive. We use decision trees with a depth of $100$ to obtain a highly expressive function class. 

As shown in Fig. \ref{fig:var}(b), RC and HC both converge to the true supervisor at the same rate.  This suggests that when model is more expressive and has enough data, it is no longer as beneficial to obtain demonstrations with RC.

\noindent \textbf{Noisy Supervisor:}
Real supervisors will not be perfect. Thus, we study the effects of noise on the performance of the two sampling techniques. This is particularly important for HC, because an expressive learner might need many more demonstrations when the supervision is noisy, perhaps making samples on the current learned policy more useful. Here we consider the case where noise is applied to the observed label, thus the robot receives control labels that are  a randomly sampled with probability $0.3$.

We use the same decision trees with depth $100$. Due to the larger expressiveness of a decision tree it is more susceptible to over-fitting.  We then compare the performance of HC vs. RC. As shown, both methods are able to converge to the expected supervisor's normalized performance, suggesting they can average out the label noise with enough data. Although it takes more data than without noise, both methods (HC and RC) converge at a similar rate to the true expected supervisor. 

\subsection{Algorithmic Convergence}
In this experiment, we show an example in which HC converges to the supervisor's performance and RC does not. We construct an environment that is representative of our insight that in real problems the policy on \emph{all} states (including recovery behaviors) is more complex than the policy on the supervisor's distribution.

We consider the example where a robot needs to learn to get to a location in a 2D, continuous domain. The robot is represented as a point mass with discrete time double integrator and linear dynamics. 

The environment contains two sets of dynamics (1 and 2): 
$$\bx_{t+1} = A\bx_{t}+B_1\bu_t+w$$
$$\bx_{t+1} = A\bx_{t}+B_2\bu_t + w$$
where $w\sim \mathcal{N}(0,0.1 I)$. The state of the robot is $\bx = (x,y,v_x,v_y)$, or the coordinates and the velocity. The control inputs to the robot $\bu = (f_x,f_y)$, or the forces in the coordinate directions. The matrix $A$ is a $4\times4$ identity matrix plus the two values necessary to update the $x,y$ state by the velocity with a timestep of $1$. $B_1$ and $B_2$, correspond to a $4\times 2$ matrix that updates only the velocity for each axis independently. This update corresponds to $\mathbf{v}_{t+1} = \mathbf{v}_t+\frac{1}{m} \mathbf{f}_t$, where $m$ is the mass of the robot.

\begin{figure}
\centering
\includegraphics{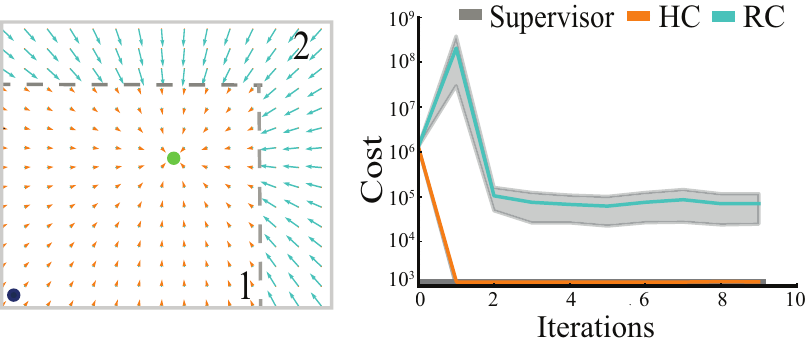}
\caption{
    \footnotesize
Left: A 2D workspace where a point mass robot is taught to go to the green circle starting from the black circle. The world is divided into to two regions with different dynamics. The supervisor is computed via infinite horizon LQG for each region, which results in two different linear matrices in region 1 and 2. Right: RC fails to converge because it is attempting to learn a policy across the two regions, whereas HC remains in region 1 and converges to the supervisor performance. The error bars shown are standard error on the mean. }
\vspace*{-20pt}
\label{fig:p_mass}
\end{figure}

The dynamics for region 1 correspond to the point robot having a mass of $m=1$. In region 2, the point robot has a larger mass of $m=4$. A target goal state lies at the origin $[0,0]$ and the robot starts out at the point $[-15,-10]$ with a velocity of zero. The boundary between regions 1 and 2 lies at $x=12$ and $y=12$, where region 2 exists when $x > 12$ and $y>12$. An illustration is shown in Fig. \ref{fig:p_mass}. The time horizon for the task is $T=35$. 

The supervisor is a switching linear system, where each linear model is computed via the infinite horizon LQG for the specified dynamics. The robot receives feedback from the supervisor based on it's current region. The robot's policy, $\pi_{\theta}$, is a linear function which we find via least squares. 

We run HC and RC in this setting and plot the performance in Fig. \ref{fig:p_mass}, averaged over $200$ trials.  HC is able to converge to the true supervisor's policy.
However, the RC approach forces the robot to enter region 2.
This prevents it from converging because it attempts to fit a single linear policy to two linear supervisors.

 \subsection{Real-World Problem}
\begin{figure}
\centering
\includegraphics{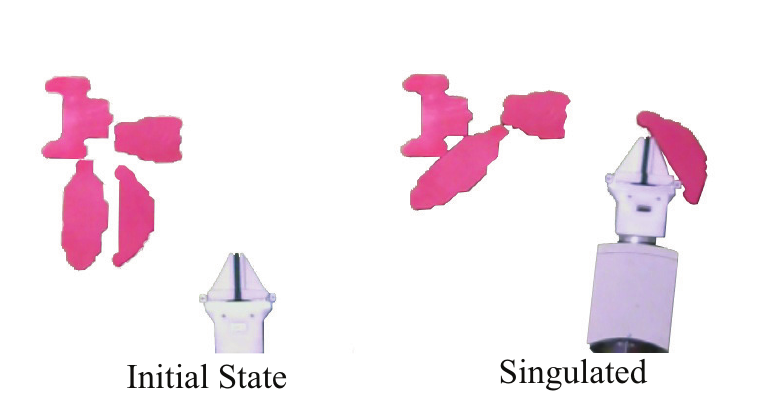}
\caption{
    \footnotesize
Left: An example initial state observed by the robot. The initial state can vary the relative position of the objects and pose of the pile. Right: A human is asked to singulate the object. Singulation is to have the robot learn to push one object away from its neighbors. A successful singulation means at least one object has its center located 10 cm or more from all other object centers.   }

\label{fig:izzy_sing}
\end{figure}

 Next, we perform a pilot user study on a real robot to test performance in practice.
Participants teach the robot to perform a singulation task (i.e., separate an object from its neighbors), illustrated in Fig. \ref{fig:izzy_sing}. A successful singulation means at least one object has its center located 10 cm or more from all other object centers. 

We hypothesize that HC will match the supervisor more accurately than RC in practice, because: 1)
RC sampling will cause the robot to try and learn more
complex behavior, 2) participants will struggle with providing
retroactive feedback.

The robot has a two-dimensional control space, $\mathcal{U}$, that consits of base rotation and arm extension. The state space of the environment, $\mathcal{X}$, is captured by an overhead Logitech C270 camera, which is positioned to capture the workspace that contains all cluttered objects and the robot arm.  The objects are red extruded polygons. We consider objects made of Medium Density Fiberboard with an average 4" diameter and 3" in height.  We use the current image of the environment as the state representation, which captures positional information. The policy is a deep neural network with the architecture from~\cite{laskeyrobot}. The network is trained using TensorFlow~\cite{tensor} on a Tesla K40 GPU. 

\begin{figure}
\centering
\includegraphics{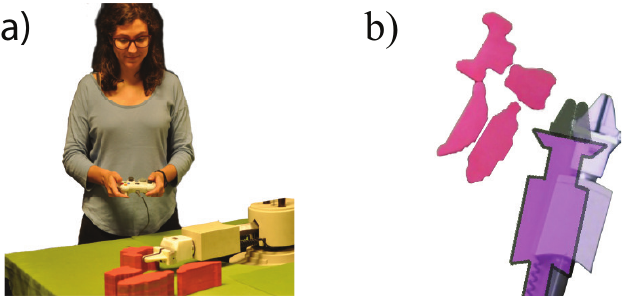}
\caption{
    \footnotesize
Two ways to provide feedback to the robot. a) In HC sampling, the human teleoperates the robot and performs the desired task. For the singulation task, the human supervisor used an Xbox Controller. b) In RC sampling, the human observes a video of the robot's policy executing and applies retroactive feedback detailing what the robot should have done. In the image shown, the person is telling the robot to go backward towards the cluster.  }
\vspace{-20pt}
\label{fig:labeling}
\end{figure}

The robot is moved via positional control implemented with PID. Similar to \cite{laskey2016shiv}, the control space $\mathcal{U}$ consists of bounded changes in rotation and translation. The control signals for each degree of freedom are continuous values with the following ranges: base rotation, $[-1.5^\circ,1.5^\circ]$, arm extension $[-1cm,1cm]$.

During training and testing the initial state distribution,  $p(\bx_0)$ consisted of sampling the translation of the cluster from a multivariate isotropic Gaussian with variance of $20$cm and the rotation was selected uniformly from the range $[-15^\circ,15^\circ]$. The relative position of the 4 objects are chosen randomly. To help a human operator place objects in the correct pose, we used a virtual overlay on the webcam feed.   

We selected 10 UC Berkeley students as human subjects. The subjects were familiar with robotics, but not the learning algorithms being assessed. They first watched a trained robot perform the task successfully. They then practiced providing feedback through RC sampling for 5 demonstrations and HC sampling for 5 demonstrations. Next, each subject performed the first 20 demonstrations via HC sampling, then performed 40 HC demonstrations and 40 RC demonstrations in a counter-balanced order. In RC, we chose $K=2$ iterations of 20 demonstrations each. The experiment took 2 hours per person on average.

In HC, we asked participants to provide 60 demonstrations to the robot using an Xbox Controller, as shown in Fig. \ref{fig:labeling}a. In RC, participants provided 20 initial demonstrations via the Xbox Controller, and then provided retroactive feedback for $K=2$ iterations of 20 demonstrations each.

Retroactive feedback was provided through a labeling interface similar to our previous work \cite{laskeyrobot} illustrated in Fig \ref{fig:labeling}b. In this interface, we showed a video at half speed of the robot's rollout to the participant. They then use a mouse to provide feedback in the form of translation and rotation. We provide a virtual overlay so that they can visualize the magnitude of their given control. A video that illustrates this setup and describes the different sampling approaches can be found at \url{https://berkeleyautomation.github.io/lfd_icra2017/}.

\begin{figure}
\centering
\includegraphics{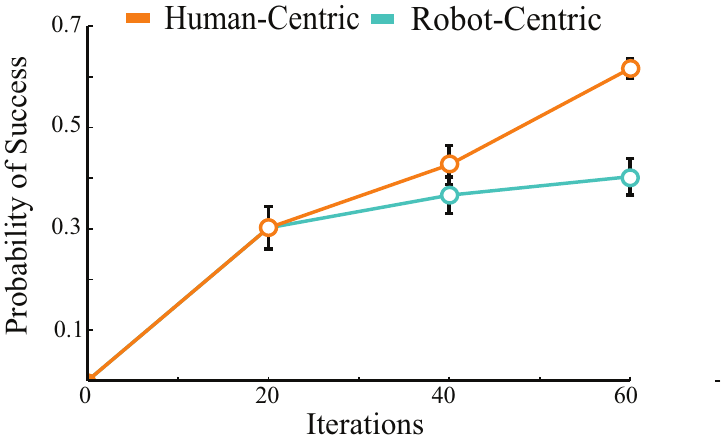}
\caption{
    \footnotesize
Average success at the singulation task over the 10 human subjects as a function of number of demonstrations. Each policy is evaluated 30 times on the a held out set of test configurations. The first 20 rollouts are from the supervisor rolling out there policy and the next 40 are collected via retro-active feedback for RC and tele-operated demonstrations for HC. HC LfD shows a 20$\%$ improvement in success at the end. The error bars shown are standard error on the mean.  }
\vspace*{-8pt}

\label{fig:izzy_rw}
\end{figure}

In Fig. \ref{fig:izzy_rw} , we show the average performance of the policies trained with RC and HC LfD. Each policy is evaluated on a holdout set of 30 initial states sampled from the same distribution as training.
The policies learned with RC have approximately $40\%$ probability of success versus $60\%$ for HC.
This suggests that HC may outperform RC when supervision is provided through actual human demonstrations.

\subsection{Understanding RC Performance in Practice}
To better understand why RC performed worse than HC, we looked for the two causes in our hypothesis: policy complexity, and difficulty of human supervision.

\noindent\textbf{Policy Complexity: }We first analyzed the complexity of behaviors collected with RC sampling. In Fig. \ref{fig:teaser}c, we show trajectories collected on a single initial state during the study. As illustrated, HC sampling is concentrated around paths needed to singulate the bottom right object. However, RC sampling places the robot in a wide variety of states, some of which would require the robot to learn how to move backward or rapidly change direction to recover. 

To better analyze this, we examined the surrogate loss on a test set of 10 randomly selected trajectories from each supervisor's dataset.  As shown in Table 1, we observed the average test error over the policies trained with 60 demonstrations in both degrees of freedom (i.e., translation and rotation) is significantly higher.
This indicates that on average the RC policies had a harder time generalizing to unseen labels in their aggregate dataset, which may be due to the complexity of the corrective actions.

\begin{table}[t]
\centering
\begin{tabular}{ R{1.75cm}||R{2.5cm}| R{2.5cm}}
 & \multicolumn{2}{c}{Surrogate Loss on Test Set} \\
 \hline
\specialcell{\bf Algorithm\\ \bf Type} & \specialcell{\bf Translation \bf (mm)} & \specialcell{\bf Rotation \\ \bf (rad)} \\
 \hline
HC LfD & $2.1\pm 0.2$ & $0.009 \pm 0.001$ \\
RC LfD & $3.4 \pm 1.0$    & $0.014 \pm 0.003$ \\
\end{tabular}
   \caption { \footnotesize  The average surrogate loss on a held out set of 10 demonstrations from the the total 60 demonstrations collected from each 10 participants. The confidence intervals are standard error on the mean, which suggest that RC LfD obtains a statistically significant higher surrogate loss in both degrees of freedom, forward and rotation. 
   }
		\tablabel{opt-p-comparison}
\vspace*{-10pt}
\end{table}

\begin{figure}
\centering
\includegraphics{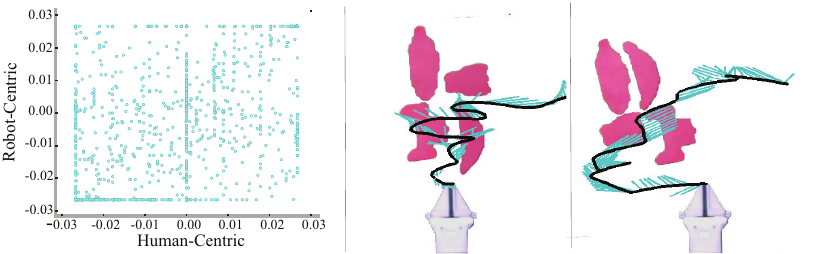}
\caption{
    \footnotesize
    Results from the post analysis examining how well retroactive feedback matched teleoperation. The scatter plot shows the normalized angle of the control applied for both HC (teleoperation) and RC (retroactive).  The large dispersion in the graph indicates that the five participants had a difficult time matching their retroactive and teleoperated controls. Two example trajectories are also shown.  The black line indicates the path from teleoperation and the teal line is the direction and scaled magnitude of the feedback given. If they matched perfectly, the teal line would be tangent to the path. 
}
\vspace*{-15pt}
\label{fig:izzy_traj}
\end{figure}

\noindent\textbf{Human Supervision. }We next hypothesized that the participants could have had trouble providing retroactive feedback consistent with their true policy. To test this, we asked 5 of the participants to provide 5 demonstrations via teleoperation. We then asked them to match their controls via the RC labeling interface. 

We measured the correlation between the controls applied via retroactive feedback and teleoperation. When calculated over all the participants and trajectories, the Pearson Correlation Coefficient in rotation and translation was 0.60 and 0.22, respectively. A smaller correlation coefficient suggests that it is harder for people to match their control in teleoperation. 

In Fig. \ref{fig:izzy_traj} we plot the control angle from HC sampling versus the control angle from RC sampling, showing that the two are not correlated. We also show two trajectories that a participant teleoperated with their retroactive labels overlayed, both of which suggest disagreement between the teleoperated and retroactive controls. Overall, our analysis suggests that RC sampling can lose the intent of the human supervisor.

\section{Theoretical Analysis} 
To better understand the empirical results from above, we contribute new theoretical analysis on RC and HC LfD. In this section, we first show the existence of environments where HC converges to the supervisor but RC does not. Then we present a new analysis of the accumulation of error for \ns.

\subsection{Algorithm Consistency}\label{sec:consistent}
In this section we work towards a fuller understanding of the tradeoffs between HC and RC. The analysis of RC LfD in \cite{ross2010reduction} is performed in the context of the regret framework of online learning. 
In online learning, at each iteration $k \in \{1,\dots,N\}$ the algorithm chooses an action $\theta_k \in \Theta$ (e.g. a policy to roll out)~\cite{shalev2011online}, and observes some loss $f_k(\theta_k)$. 
Online learning is analyzed in the context of a regret guarantee: an upper bound (e.g. $\sqrt{N}$) on the cumulative loss incurred by the algorithm relative to taking just the best single action in $\Theta$:
\begin{align*}
\sup_{\theta \in \Theta} \sum_{k=1}^N f_k(\theta_k) - f_k(\theta).
\end{align*}
In the context of robotics and specifically in the case of RC LfD, taking an action $\theta_k$ at iteration $k$ is rolling out a policy dictated by $\theta_k$ which induces a series of states $\bx_{k,1},\dots,\bx_{k,T}$ and taking the aggregate loss evaluated with respect to {\em these} states.  
Because these particular states could have been produced by a poor policy due to initialization, they could have little relevance to the task of interest and consequently, it makes little sense to compare to the policy that performs best on these particular states.
What is important is the absolute performance of a policy on the task of interest. 
In the notation of Ross et al.~\cite{ross2010reduction} this notion of relative regret is encoded in the $\epsilon_N$ error term that appears in their bounds and is defined relative to the particular set of rollouts observed by the RC LfD policy.

As an example for why low-regret may be uninformative, consider a car-racing game where the car has constant speed and the policy only chooses between straight, left, or right actions at the current state. 
Suppose at some point in the race the car encounters a ``v'' in the road where one path is a well-paved road and the other path is a muddy dirt road with obstacles. 
The car will finish the race faster by taking the paved road, but due either to the stochastic dynamics or imperfectness of the supervisor it is possible that after just a small number of supervisor demonstrations given to the robot to initialize RC LfD, a poor initial policy will be learned that leads the car down the dirt road instead of the paved road. 
When this policy is rolled out the supervisor will penalize the decision at the time point of taking the dirt versus paved road. 
But if the policy's actions agree with the supervisor once the car is on the dirt road (i.e., making the best of a bad situation) this policy will incur low-regret because the majority of the time the policy was acting in accordance with the supervisor.
Thus, in this example a globally bad policy will be learned (because it took the dirt road instead of the paved road) but relative to the best policy acting on the dirt road, it performs pretty well. 

The next theorem shows an instance in which an initial policy can be different enough from the optimal policy that the states visited by the initial policy -- even with corrective feedback from the supervisor --  are not informative or relevant enough to the task to guide RC LfD to the optimal policy. 

\begin{theorem}
For a given environment (state space, action space, dynamics mapping state-action pairs to states, and a loss function) and policy class $\Theta$,
let $\theta_{HC}^N$ be the policy learned from $N$ supervisor demonstrations,
and let $\theta_{RC}^N$ be the policy learned by the $RC$ procedure described in Section~\ref{sec:PS}, initialized with $m$ supervisor demonstrations.  
Then there exists an environment and policy class $\Theta$ such that 
\begin{itemize}
\item $\theta^*$ is the unique minimizer of $E_{p(\tau \mid \theta)}[ J( \theta ) ]$,
\item $\lim_{N \rightarrow \infty} \theta_{HC}^N = \theta^*$ with probability $1$,
\item $\lim_{N \rightarrow \infty} \theta_{RC}^N \neq \theta^*$ with probability at least $c e^{-m}$
\end{itemize}
for some universal constant $c$.
In other words, even with infinite data RC may converge to a suboptimal policy while HC converges to the best policy in the class that uniquely achieves the lowest loss.
\end{theorem}

\begin{figure}
\centering
\includegraphics{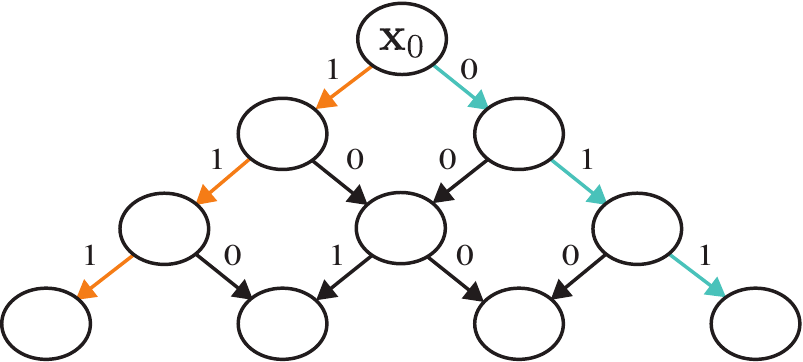}
\caption{
    \footnotesize
A Directed Acyclic Graph, where a robot is being taught by a supervisor to descend down and achieve maximum cumulative reward, which is shown via the numbers on each branch. Each node has the action the supervisor would select, either left or right, $\lbrace L, R \rbrace$. The HC method converges to the Orange path, which is optimal. However, the RC method converges to the Teal path, because it tries to learn on examples from that side of the tree.}
\vspace*{-20pt}
\label{fig:c_ex}
\end{figure}

\begin{proof}
Let $\{ \{ (\bx_{k,t},\bu_{k,t} \}_{t=0}^3 \}_{k=1}^N$ denote $N$ trajectories.
Consider an environment with a deterministic initial state at the root of the DAG of Figure~\ref{fig:c_ex} so that $p(\bx_{k,0} = \texttt{root}) = 1$ for all $k$. 
The space of policies are constant functions $\Theta = \{L,R\}$ where if $\theta=L$, then regardless of the state, the control input $\bu_{k,t}$ will be to take the left child (and analogously for $\theta=R$ taking the right child). 
For any state in the DAG with children, let $\phi(\bx_{k,t},\theta)$ denote the left child of $\bx_{k,t}$ if $\theta=L$. Otherwise, denote it the right child. 
The dynamics are described as follows: for some $\mu \in (0,1/4]$ to be defined later, if $\theta=L$ and $\bx_{k,t}=\texttt{root}$ then $p(\bx_{k,1}=\phi(\bx_{k,0},R))=\mu$ and $p(\bx_{k,1}=\phi(\bx_{k,0},L))=1-\mu$, but if $\theta=R$ then the right child is chosen with probability $1$. If $\bx_{k,t} \neq \texttt{root}$ then $\bx_{k,t+1} = \phi(\bx_{k,t},\theta)$.

Assume that the supervisor $\pi^*: \mathcal{X} \rightarrow \{L,R\}$ acts greedily according to the given rewards given in the DAG so that $\pi^*(\bx_{k,t})=L$ if the reward of the left child exceeds the right child and $\pi^*(\bx_{k,t})=R$ otherwise.
Finally, define the state loss function $\ell(\cdot,\cdot)$ as the $0/1$ loss so that after $N$ trajectories, the loss is given as $J_N(\theta)=\sum_{k=1}^N \sum_{t=0}^2 \mathbf{1}\{ \pi^*(\bx_{k,t}) \neq \theta \}$ for all $\theta \in \{L,R\}$. Note that $\widehat{\theta}_N = \arg\min_{\theta \in \{L,R\}} J_N(\theta)$ is equivalent to looking at all actions by the supervisor over the states and taking the majority vote of $L$ versus $R$ (i.e., the states in which these actions are taken has no impact on the minimizer). Note that we have not yet specified how the states $\bx_{k,t}$ were generated. 

We can compute the true loss when the trajectories are generated by $\theta \in \{L,R\}$. Let the empirical distribution of observed states under a fixed action $\theta \in \{L,R\}$ be given by $p(\tau \mid \theta)$, then
\begin{align*}
 E_{p(\tau \mid \theta=L) } J(\theta=L) &= p( \bx_{k,1} = \phi(\bx_{k,0},L) ) \cdot 0 +\\
 & p( \bx_{k,1} = \phi(\bx_{k,0},R) ) \cdot 2  = 2 \mu\\
   E_{p(\tau \mid \theta=R) } J(\theta=R) &= 1
\end{align*}
which implies that $\theta_* = L$ and performs strictly better than $R$ whenever $\mu < 1/2$.

It follows by the stochastic dynamics that the demonstrated action sequence by the supervisor equals $\{L,R,R\}$ with probability $\mu$ and $\{L,L,L\}$ with probability $1-\mu$.
After $m$ supervisor sequences, the expected number of $L$ actions is equal to $\mu+3(1-\mu)=3-2\mu$ while the expected number of $R$ actions is equal to just $2\mu$. 
By the law of large numbers, if only supervisor sequences are given then $\arg\min_{\theta \in \{L,R\}} J_m(\theta) \rightarrow L = \theta^*$ as $m \rightarrow \infty$ since $3-2\mu>2\mu$ for all $\mu<1/2$. 
In other words, $\theta_{HC}^N \rightarrow \theta^*$ as $N \rightarrow \infty$.

We now turn our attention to the $RC$ policy.
Note that if after $m$ supervisor action sequences we have that the number of observed $R$'s exceeds the number of observed $L$'s, then the $RC$ policy will define $\theta_{RC}^m = R$.  This proof assumes $\beta=0$ in the RC (DAgger) algorithm~\cite{ross2010reduction}, however it can be extended to include $\beta$ without loss of generality. 
It is easy to see that a policy rolled out with $\theta=R$ will receive the supervisor's action sequence $\{L,R,R\}$ and thus $R$ will remain the majority vote and consequently $\theta_{RC}^N=R$ for all $N \geq m$.
What remains is to lower bound the probability that given $m$ supervisor demonstrations, $\theta_{RC}^m = R$. 

For $k=1,\dots,m$ let $Z_k \in \{0,1\}$ be independent Bernoulli($\mu$) random variables where $Z_k=1$ represents observing the supervisor sequence $\{L,R,R\}$ and $Z_k=0$ represents observing $\{L,L,L\}$.
Given $m$ supervisor sequences, note that the event $\arg\min_{\theta \in \{L,R\}} J_m(\theta) = R$ occurs if $\frac{1}{m}\sum_{k=1}^m Z_k > 3/4$. 
Noting that $\sum_{k=1}^m Z_k$ is a binomial($m$,$\mu$) random variable, the probability of this event is equal to 
\begin{align*}
\sum_{k=\lfloor 3m/4 \rfloor+1}^m \binom{m}{k} \mu^k (1-\mu)^{m-k} \geq \Phi\left( \frac{\lfloor 3m/4 \rfloor+1 - \mu m}{ \sqrt{m \mu(1-\mu)}} \right) 
\end{align*} 
where we have used Slud's inequality \cite{slud1977} to lower bound a binomial tail by a Gaussian tail. 
Setting $\mu=1/4$ we can further lower bound this probability by $\Phi\left( \frac{m + 2}{ \sqrt{3m/4}} \right) \geq c e^{-m}$ for some universal constant $c$.
Consequently, with probability at least $c e^{-m}$ we have that $\lim_{N \rightarrow \infty} \theta_{RC}^N = R \neq \theta^*$ which implies that the expected loss of $\lim_{N \rightarrow \infty} \theta_{RC}^N$ must exceed the expected loss of $\theta_*=L$. 
\end{proof}

We note there exist techniques to correct for this problem.
One approach is to consider the value of each action and
select actions that lead to higher reward during roll-out~\cite{ross2014reinforcement}.
However, this assumes the robot has access to a reward
function that provides sufficient information during each roll-out. Another solution is to
increase the model expressiveness of the robot’s policy class.
However, this could require more data than HC methods~\cite{vapnik1992principles}.

\section{Discussion and Future Work}

Motivated by recent advances in Deep Learning from
Demonstrations for robot control, this paper reconsiders HC
and RC sampling in the context of human supervision, finding
that policies learned with HC sampling perform equally well
or better with classes of highly-expressive learning models and
can avoid some of the drawbacks of RC sampling. We further
provide new theoretical contributions on the performance of
both RC and HC methods.

It is important to note that there are challenges to using
highly- expressive policies. When training complex models
such as Recurrent Neural Networks or Differentiable Neural
Computers, it may be difficult to achieve low training error or enough data~\cite{bengio2015scheduled,graves2016hybrid}. Thus, future work is to consider how to best learn robust policies that are Human-Centric.

\section{Acknowledgments}
This research was performed at the AUTOLAB at UC
Berkeley in affiliation with the UC Berkeley’s Algorithms, Machines,
and People Lab, BAIR, BDD, and the CITRIS ”People
and Robots” (CPAR) Initiative: \url{http://robotics.
citris-uc.org}. KJ is supported by ONR awards N00014-
15-1-2620 and N00014-13-1-0129. JM is the Department of
Defense (DoD) through the National Defense Science \&
Engineering Graduate Fellowship (NDSEG) Program. We
acknowledged David Gealy and Menglong Guo for their help
with the hardware setup . We also thank Florian Pokorny,
Drew Bagnell, Hal Duame III and Roy Fox for valuable insight

\bibliographystyle{IEEEtranS}
\bibliography{references}

\end{document}